\documentclass[manuscript]{clv2025}

\usepackage{fontspec}
\usepackage{latexsym}
\usepackage{microtype}
\usepackage{graphicx}
\usepackage{booktabs}
\usepackage{amsmath}
\usepackage{xcolor}
\usepackage{enumitem}

\usepackage{polyglossia}
\setmainlanguage{english}
\setotherlanguage[variant=ancient]{greek}
\newfontfamily\greekfont{FreeSerifb}[
  Extension = .otf,
  UprightFont = FreeSerifb,
  ItalicFont = FreeSerifbItalic,
  BoldFont = FreeSerifbBold,
]

\jvol{vv}
\jnum{nn}
\jyear{2026}
\jname{}
\jinfo{}
 \renewcommand{\footmark}{}   

\title{`The Order in the Horse's Heart': A Case Study in LLM-Assisted Stylometry for the Discovery of Biblical Allusion in Modern Literary Fiction}

\author{Ewan Cameron\thanks{\hspace{0.5em}ORCID: \texttt{0000-0002-8842-3811}. Google Scholar: \texttt{https://scholar.google.com/citations?user=KQVMIHAAAAAJ}}}

\affilblock{%
\affil{Actuarial Mathematics \& Statistics, School of Mathematical \& Computer Sciences, Heriot-Watt University, Edinburgh EH14 4AS, UK\\
\texttt{Ewan.Cameron@hw.ac.uk}}%
}

\runningtitle{`The Order in the Horse's Heart': LLM-Assisted Discovery of Biblical Allusion}
\runningauthor{Cameron}

\begin{document}
\maketitle

\begin{abstract}
We present a dual-track pipeline for detecting biblical allusions in literary fiction and apply it to the novels of Cormac McCarthy. A bottom-up embedding track uses inverse document frequency to identify rare vocabulary shared with the King James Bible, embeds occurrences in their local context for sense disambiguation, and passes candidate passage pairs through cascaded LLM review. A top-down register track asks an LLM to read McCarthy's prose undirected to any specific biblical passage for comparison, catching allusions not distinguished by word or phrase rarity. Both tracks are cross-validated by a long-context model that holds entire novels alongside the KJV in a single pass, and every finding is checked against published scholarship. Restricting attention to allusions that carry a textual echo---shared phrasing, reworked vocabulary, or transplanted cadence---and distinguishing literary allusions proper from signposted biblical references (similes naming biblical figures, characters overtly citing scripture), the pipeline surfaces 349 allusions across the corpus. Among a target set of 115 previously documented allusions retrieved through human review of the academic literature, the pipeline independently recovers 62 (54\% recall), with recall varying by connection type from 30\% (transformed imagery) to 80\% (register collisions). We contextualise these results with respect to the value-add from LLMs as assistants to mechanical stylometric analyses, and their potential to facilitate the statistical study of intertextuality in massive literary corpora.
\end{abstract}

\bigskip
\section{Introduction}
\label{sec:intro}

Allusion---the creative borrowing by one writer from the work of another---may well be the foremost structural device by which authors across the ages have responded to the conscious burden of their literary inheritance, thereby acknowledging ``the predicaments and responsibilities'' so placed upon them \citep{ricks-2002-allusion}.  As far back as the Archaic era of Ancient Greece we have examples of sublime allusive constructions that depend not only on the reader's familiarity with an established literary and creative canon but with the nature and conventions of allusion itself.  In an apparent reworking of ``\textgreek{ὅσον τ' ἐπικίδναται ἠώς}'' (\textit{Iliad}~7.451) as ``\textgreek{ὅσον τ’ ἐπὶ γῆν κίδναται ἠέλιος}'' (fr.~2.8), Mimnermus must anticipate that the astute reader's amongst his audience will draw the connection despite the structural inversion of meaning \citep{griffith1975man, dover1971theocritus}, identifying the lexical similarity as deliberate rather than accidental.   Yet as Tennyson reminds us,  ``No man can write a single passage to which a parallel one may not be found somewhere in the literature of the world'' \citep{tennyson-1897-memoir}.  The sheer volume of humanity's creative output effectively dictates that the scholarly study of literary allusion can benefit almost as much from historical research into the archival notes and personal library of an author \citep{crews-2017-books} as from literary theory or stylometric analysis. 

There is perhaps no weightier tome of cultural inheritance for Western literature than the Bible itself \citep{frye-1982-great-code}.  The phrasing and moral vocabulary of the Bible permeate everyday speech and its stories and characters serve as archetypes across diverse genres of secular creative writing \citep{crystal-2010-begat}.  Few modern novelists have drawn on this inheritance more deliberately than Cormac McCarthy, whose writing absorbs and repurposes scriptural language at every level from direct quotation to prophetic register and parataxis to textual and narrative allusion \citep{alter-2010-pen, lincoln-2009-mccarthy}.  By way of example for those unfamiliar with McCarthy's writing, his second novel, \textit{Outer Dark} (1968; the title itself derived from the Gospel of Matthew), features a literal herd of Gadarene swine \citep{schafer1977hard,cant-2008-mccarthy}, and the last paragraph of his Pulitzer Prize-winning novel, \textit{The Road} (2006), reworks Genesis 2:7 as ``the breath of God was his breath yet though it pass from man to man through all of time'' \citep{alter-2010-pen}.  

The aim of this manuscript is to explore the potential of LLM-assisted stylography for automatic detection of allusion and intertextuality at scale.  If LLMs are able to deliver the contextual reasoning needed to transform mechanical stylometric metrics of similarity into effective and efficient probes of allusive writing, a new quantitative science of literature may well be opened before us.  A science that could measure reliably and reproducibly the relative strength of influence of literary inheritance on writing through the ages, of its relation with genre, movement, nationality and language, and of the technical devices by which it is accomplished.  By no means could one rationally propose that in being automated and quantitative in character such a science would hold claim to a higher truth than (human) close reading and traditional means of literary study \citep{hacking-2012-styles}.  Yet it would be surprising, on the other hand, if some insights were not more readily, if not solely, attainable through the sheer scale at which computational analyses can be conducted.  We take the detection of biblical allusion in the novels of Cormac McCarthy as our case study---not because we anticipate a wealth of insights into McCarthy's biblical influence yet to be gleaned beyond those already established through traditional scholarship, but because the richness of that scholarship provides a ground truth against which to calibrate the pipeline.

Computational stylometry has a rich history of application to intertextual analyses.  Inverse document frequency (IDF) tables have been used to weight or filter word, stem and n-gram matches by lexical frequency \citep{buechler-2012-recall, smith-etal-2014-detecting}, and contextual embeddings deployed towards fuzzy phrase matching \citep{burns-etal-2021-profiling} and uncovering thematic similarities \citep{li-2024-genealogies, barre-2024-latent}.  Yet with the sheer diversity of allusive styles and the difficulty of thereby achieving high sensitivity and specificity through mechanical features alone, the inevitability of extensive post-pipeline human review has remained a rate-limiting step for scaling these methods.  Although the extent to which they deliver `reasoning' remains under theoretical debate \citep{wu-etal-2024-reasoning}, there is emerging empirical evidence that LLMs can act effectively as proxies for human review in such settings.  \citet{yang-etal-2025-interideas} combined LLMs with retrieval-augmented generation to extract philosophical references at scale, while \citet{umphrey-etal-2024-expert} showed that expert-in-the-loop LLM workflows can identify intertextual connections in biblical Greek texts.  In this manuscript we continue this exploration of the potential of LLM-assisted stylometric detection of allusion, developing a sophisticated pipeline for automated discovery and extensively comparing its performance against human scholarship.

Our pipeline combines two complementary detection modalities. A \emph{bottom-up embedding track} uses IDF to identify rare vocabulary and phrasing shared between McCarthy and the King James Version (KJV) of the Bible, embeds each occurrence in its local context for sense disambiguation, and passes candidate passage pairs through cascaded LLM review. And, a \emph{top-down register track} asks an LLM to read McCarthy's prose---undirected to any specific biblical passage for comparison---and flag passages exhibiting biblical cadence and imagery, catching allusions that use only common words. Both tracks are cross-validated by a long-context model that holds entire novels alongside large portions of the KJV in a single pass, and every finding is checked against published scholarship and the Semantic Scholar API.

Our contributions are:
\begin{enumerate}[nosep,leftmargin=*]
    \item A \textbf{provenance-tagged catalogue} of 349 biblical allusions across all twelve McCarthy novels---234 of them previously unrecorded---with every finding traceable to its discovery track and verifiable against quoted source text.
    \item A \textbf{dual-track architecture} whose embedding and register components approach the same allusions from independent starting points, providing built-in convergence checks.
    \item A \textbf{stratified error analysis} that localises the pipeline's 54\% recall ceiling to the mechanical retrieval stages rather than to LLM judgement.
    \item A \textbf{Bayesian framework} for contextualising pipeline performance: treating allusion recovery as an imperfect observation process whose stratum-specific sensitivity and specificity can be estimated from calibration data, with adaptive design to direct both computational and human effort where uncertainty is greatest.
    \item A \textbf{contamination probe} using LLM-written McCarthy pastiche with planted KJV echoes, demonstrating that frontier LLMs perform cross-textual analysis rather than recalling memorised scholarship.
    \item An \textbf{end-to-end cost} of \$50--80 in API calls on consumer hardware, making the methodology accessible as a reusable template for other author--source pairings.
\end{enumerate}

\bigskip
\section{Related Work}
\label{sec:related}

\paragraph{Allusion versus text reuse}
The computational methods surveyed in \S\ref{sec:intro} have been developed overwhelmingly for the detection of \emph{text reuse}: near-verbatim borrowing whose surface traces---shared n-grams, parallel phrase structure, embedding proximity---are most amenable to mechanical scoring.  Literary allusion is a fundamentally different phenomenon.  Where text reuse asks ``do these passages share words?'', allusion asks ``does one passage creatively respond to another?''---a question that may be answered in the affirmative even when the two texts share no vocabulary at all, and in the negative even when they share a great deal.  \citet{bamman-etal-2024-classification}, benchmarking LLMs against supervised baselines on ten classification tasks in cultural analytics, find that performance varies sharply with the nature of the literary question posed; allusion detection sits at the hard end of this spectrum, since it demands comparative judgement across two texts rather than classification of one.  Our architecture is designed for this harder problem, and our evaluation is calibrated accordingly.

\paragraph{McCarthy's allusive practice}
McCarthy is an unusually instructive test case for allusion detection because his borrowings are at once pervasive and technically diverse.  \citet{crews-2017-books}, working from the annotated volumes in McCarthy's personal library held at the Wittliff Collections at Texas State University, highlights over a hundred works with strong evidence as direct literary influences on his writing and takes as its title McCarthy's own dictum that ``books are made out of books.''  What makes these borrowings difficult for any single detection method is the range of transformations they undergo.  Cummings's ``in Just- / spring'' and ``goat-footed balloonMan'' are decomposed in \textit{Suttree} into ``In just spring the goatman came''---a compound allusion that scatters its source across a new syntactic frame.  Auden's poignant elegiac verse in remembrance of Yeats (``What instruments we have agree / The day of his death was a dark cold day'') receives a parodic reworking in the same novel as the motley crew crudely discuss the demise of Old Cecil (``All agreed that the day of his death was a cold one'')---an allusion anchored by such an innocuous phrase that its confident identification require the contextual recognition of the funeral scene and the comedic inversion of tone.  The diversity of McCarthy's technique thus provides a natural stress test for any architecture that claims generality.

\paragraph{McCarthy and the Bible}
The biblical stratum of McCarthy's writing has been a sustained concern of literary criticism since at least \citet{schafer1977hard}'s early reading of \textit{Outer Dark}, and it has deepened considerably with the dedicated monographs of \citet{cant-2008-mccarthy} and \citet{lincoln-2009-mccarthy}.  \citet{alter-2010-pen} devotes a chapter of \textit{Pen of Iron} to the KJV's influence on McCarthy's prose, arguing that he is among the very few modern novelists for whom the rhythms of the King James Bible remain a living compositional resource rather than a merely inherited idiom.  More specialised treatments include \citet{potts-2015-sacrament} on sacramental theology across the full corpus, \citet{mundik-2017-bloody} on metaphysical violence and biblical cosmology in \textit{Blood Meridian}, and \citet{broncano-2014-religion} on religion in the border fiction. \citet{noble-2020-bible} surveys McCarthy's scriptural engagement across diction, allusion, typology, and exegetical scenes, and \citet{lewis-2023-word} catalogues verse-level allusions in the Appalachian novels.  Rich as this body of work is, it has accumulated piecemeal: each critic examines one or a few novels, and no single study attempts a systematic account of McCarthy's biblical borrowings across the full twelve-novel corpus.  The reference table against which we calibrate our pipeline draws on all of these sources and others, compiling notes on 218 allusions documented in the accessible scholarship, of which 124 may be classified as `textual' (featuring shared vocabularly, phrasing, syntax, cadence or transformed diction).  This represents, to our knowledge, the most comprehensive ground truth for McCarthy's biblical borrowings presently available, assembled from over thirty publications spanning five decades of criticism.  Yet it is almost certainly an undercount of what traditional scholarship has noticed.  McCarthy criticism is dispersed across monographs, edited volumes, and journal articles, and as many of these are behind paywalls a complete census would require access to several further works that we were unable to obtain.

\bigskip
\section{Method}
\label{sec:method}

\begin{figure*}[t]
  \centering
  \includegraphics[width=0.75\textwidth]{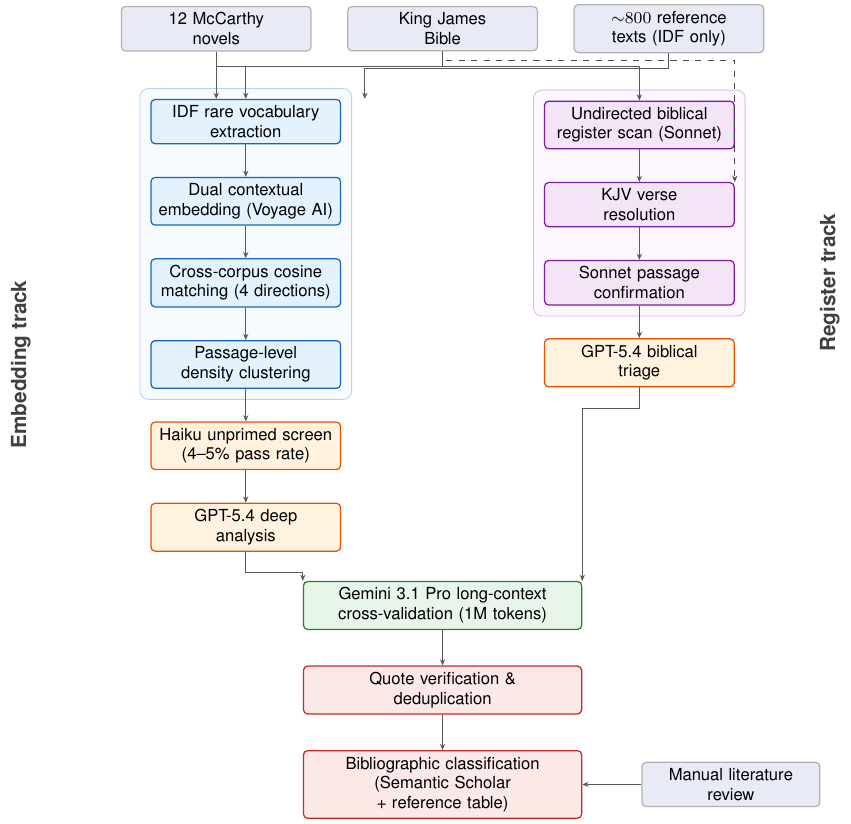}
  \caption{Pipeline architecture.  The embedding track (left, blue) surfaces allusions through shared rare vocabulary; the register track (right, purple) catches allusions that use common words in distinctively biblical patterns.  Both tracks feed into a shared verification layer (bottom).}
  \label{fig:pipeline}
\end{figure*}

The pipeline combines two complementary discovery strategies with cascaded verification and automated bibliographic checking.  A bottom-up \emph{embedding track} is driven by statistical signals in shared rare vocabulary between McCarthy's novels and the KJV; a top-down \emph{register track} is driven by LLM reading at full-novel scale for biblical cadence and phrasing that uses only common words.  Both tracks feed into a shared verification layer; Figure~\ref{fig:pipeline} provides an overview.

\bigskip
\subsection{Corpus}
\label{sec:corpus}

The target corpus comprises all twelve of Cormac McCarthy's published novels, from \textit{The Orchard Keeper} (1965) through \textit{Stella Maris} (2022), totalling approximately 3.5 million characters.  The source text against which allusions are sought is the King James Bible---the translation whose influence on McCarthy's prose style is most extensively documented in the critical literature \citep{alter-2010-pen}.  An additional reference corpus of approximately 800 texts spanning classical, early modern, and twentieth-century literature is used solely for the computation of inverse document frequency scores.

\bigskip
\subsection{Statistical Signal Extraction}
\label{sec:signal}

Inverse document frequency scores \citep{jones-1972-idf} are computed across the reference corpus to identify vocabulary that is distinctive rather than merely shared.  A word or phrase qualifies as \emph{rare} in our study if its IDF exceeds 4.0---that is, if it appears in no more than 15 of the approximately 800 reference texts---a threshold that retains distinctively biblical vocabulary such as \textit{raiment}, \textit{smote}, and \textit{dominion} while filtering out words in common use in modern English.  Candidates are further filtered to remove OCR artefacts, Spanish-language tokens, and proper names.

Each occurrence of a rare word or phrase is then embedded using two Voyage AI models: \texttt{voyage-4} for semantic sense disambiguation and \texttt{voyage-4-large} for broader contextual similarity.  The embedding is computed over a context snippet extending up to four words in either direction but truncated at clause- or sentence-breaking punctuation, rather than the bare term, producing sense-specific vectors that can distinguish, for example, \textit{still} in the sense of motionless from \textit{still} in the sense of distillation---and thereby prevent spurious cross-corpus matches between homographs.  Cosine similarity is then used to identify shared rare vocabulary across four matching directions (word--word, phrase--phrase, and the two cross-type pairings), with same-type thresholds of $\geq 0.88$ and cross-type thresholds of $\geq 0.82$, each validated by a context-similarity floor of $\geq 0.40$.  This process yields approximately 3{,}000 matched passage pairs across the twelve novels.  A subsequent density-mapping stage reranks these pairs by how densely convergent lexical evidence clusters within sliding windows at three scales.

\bigskip
\subsection{Cascaded LLM Review}
\label{sec:llm-review}

The statistical signal extraction casts a deliberately wide net; the role of LLM review is to separate genuine allusive connections from coincidental vocabulary overlap.  Every matched passage pair---a roughly 1{,}000-word McCarthy chunk alongside a roughly 1{,}000-word KJV chunk---is sent to Claude Haiku\footnote{Model: \texttt{claude-haiku-4-5-20251001}.} for unprimed assessment.  Crucially, the model is \emph{not} told which words triggered the match; it must judge the connection from the passages alone.  Each pair is rated \textbf{Strong}, \textbf{Moderate}, \textbf{Weak}, or \textbf{None}.  On the KJV portion of the candidate set (${\sim}$16{,}000 pairs), approximately 87\% are rejected as \textbf{None} on first pass.  Only pairs reaching \textbf{Strong} or \textbf{Moderate} proceed to deep analysis by GPT-5.4, which classifies each finding by connection type---\emph{verbal echo}, \emph{inverted allusion}, \emph{parodic homage}, \emph{register collision}, \emph{thematic transplantation}, \emph{morphological transformation}---and returns exact quotations from both texts.

\bigskip
\subsection{Biblical Register Scan}
\label{sec:register}

The embedding track requires shared rare vocabulary as its point of entry, and consequently misses allusions that operate through \emph{register} and \emph{cadence} rather than distinctive diction.  Some of McCarthy's most powerful biblical language uses only common English words arranged in distinctively biblical patterns: the sentence ``the last and the first suffer equally'' (\textit{Suttree}) contains no rare words yet its cadence and semantic content is unmistakably that of Mark~10:31.  A dedicated register scan addresses this gap by asking Claude Sonnet\footnote{Model: \texttt{claude-sonnet-4-6-20250514}.} to read every McCarthy passage \emph{without any KJV text supplied in the context window} and to flag biblical cadence, prophetic register, liturgical patterns, or transformed scriptural imagery wherever it finds them.  For each flagged passage, the model identifies the likely KJV source at book-and-verse specificity; suggested verses are then resolved to character offsets in the KJV text, paired with the corresponding KJV passage, and confirmed by a second Sonnet call.  Confirmed pairs undergo the same GPT-5.4 deep analysis described in \S\ref{sec:llm-review}.

\bigskip
\subsection{Long-Context Cross-Validation}
\label{sec:gemini}

The embedding track and the register scan have complementary blind spots: the former misses allusions built from common words, while the latter may miss structurally subtle connections that do not announce themselves through register.  To cross-validate both tracks we employ Gemini 3.1 Pro\footnote{Each novel requires 2--4 calls at approximately 500K--650K tokens.} with its million-token context window to hold an entire McCarthy novel alongside substantial portions of the KJV in a single pass---a fundamentally different search strategy that asks the model to \emph{read everything and find connections} rather than to \emph{match words then verify}.  Four complementary passes scan each novel: against Genesis through Job; against Psalms through Revelation; with the pipeline's matched rare words supplied as hints; and as a critical review of existing pipeline findings, flagging false positives and omissions.  The two approaches prove complementary in practice: for \textit{The Road}, for example, the embedding pipeline surfaced ``the sleep of death'' (Psalm~13:3) and specific Deuteronomic curse formulae that Gemini missed, while Gemini identified the father's sustained engagement with Job~23---an allusion that operates through theological argument rather than shared vocabulary.

\bigskip
\subsection{Verification and Classification}
\label{sec:verification}

Findings from each detection path are then deduplicated at two levels: first, collapsing multiple chunk pairings that target the same McCarthy passage by fuzzy word-overlap matching; and second, collapsing multiple GPT-5.4 classifications of the same passage.  Each allusion is checked against a manually-compiled reference table of biblical allusions documented in McCarthy scholarship and against the Semantic Scholar API, and assigned to \textbf{Attested} (documented in published scholarship) or \textbf{Unattested} (no prior citation found).  Detailed prompts and parameters are given in Appendix~\ref{sec:appendix-pipeline}; the full twelve-novel analysis costs approximately \$50--80 in API calls and runs in roughly 8--12 hours on consumer hardware, the rate-limiting step being conservative API call pacing rather than local computation.

\bigskip
\section{Results}
\label{sec:results}

The results of the pipeline are summarised in Table~\ref{tab:summary} and presented in full in the supplementary workbook (Appendix~\ref{sec:appendix-catalogue}).  The full catalogue documents 527 biblical connections across the twelve novels, but the main analysis restricts attention to the 364 that carry a \emph{textual} echo---shared phrasing, reworked vocabulary, or transplanted cadence---excluding the 155 that operate purely through structural parallels, character typology, sacramental theology, or naming conventions.  Within this textual category, 15 of the 364 findings are better described as \emph{biblical references} than \emph{allusions} in the sense that Ricks's term implies: passages where a character overtly cites scripture (``The good book says that he that lives by the sword shall perish by the sword''), uses a biblical proper noun in simile (``like the soldiers of Pharaoh''), or employs a proverbial phrase with framing that marks it as biblical (``It's a judgment.\ Wages of sin and all that'').  These are genuine engagements with the Bible, but they lack the unmarked, transformative quality that distinguishes literary allusion from cultural reference.  We retain these entries in the supplementary catalogue (flagged accordingly) but exclude them from the headline counts, leaving \textbf{349 textual allusions} proper.

Of these 349, the pipeline independently discovered 296 and 53 were attested in published scholarship but not independently recovered (\textbf{Missed}).  The 296 pipeline discoveries include 62 that are also documented in the accessible scholarship (\textbf{Attested}) and 234 that are not (\textbf{Unattested}).  Each finding is also tagged with its discovery track (Pipeline / Gemini / Pipeline+Gemini); the two tracks converge on a substantial subset of findings, approaching the same allusions from different starting points.

\paragraph{What the pipeline finds and what it misses}
Table~\ref{tab:by-type} breaks recall down by connection type (allusions only, excluding references and duplicates), where recall is the proportion of attested allusions of each type that the pipeline independently rediscovered.  Register collisions are recovered most reliably (80\%), followed by direct quotations (75\%) and inverted allusions (59\%).  Thematic transplantations (57\%), verbal echoes (50\%), and parodic homages (42\%) cluster near the overall 54\% recall rate.  Transformed imagery is the hardest category (30\%), reflecting cases where a biblical image has been reworked beyond easy lexical recognition.  The 155 non-textual allusions excluded from the main analysis are documented in the supplementary workbook; recall on these categories is very low by design, since they lack the shared vocabulary or cadence that the pipeline requires as an entry point.

\begin{table}[t]
\centering
\small
\begin{tabular}{lrrrrrr}
\toprule
\textbf{Connection type} & \textbf{Tot.} & \textbf{Att.} & \textbf{Un.} & \textbf{Fnd.} & \textbf{Miss.} & \textbf{Recall} \\
\midrule
Register collision & 22 & 5 & 17 & 21 & 1 & 80\% \\
Direct quotation & 9 & 8 & 1 & 7 & 2 & 75\% \\
Inverted allusion & 53 & 22 & 31 & 44 & 9 & 59\% \\
Thematic transplantation & 114 & 28 & 86 & 102 & 12 & 57\% \\
Verbal echo & 96 & 30 & 66 & 81 & 15 & 50\% \\
Parodic homage & 30 & 12 & 18 & 23 & 7 & 42\% \\
Transformed imagery & 25 & 10 & 15 & 18 & 7 & 30\% \\
\midrule
\textbf{Total} & \textbf{349} & \textbf{115} & \textbf{234} & \textbf{296} & \textbf{53} & \textbf{54\%} \\
\bottomrule
\end{tabular}
\caption{Recall stratified by connection type (349 allusions; 15 signposted references excluded).  \emph{Tot.}\ = total count; \emph{Att.}\ = attested in published scholarship; \emph{Un.}\ = unattested; \emph{Fnd.}\ = independently discovered by the pipeline; \emph{Miss.}\ = attested but not independently discovered; \emph{Recall} is the proportion of attested allusions independently rediscovered.}
\label{tab:by-type}
\end{table}

\begin{table}[t]
\centering
\small
\begin{tabular}{lrrrrr}
\toprule
\textbf{Novel} & \textbf{Tot.} & \textbf{Att.} & \textbf{Un.} & \textbf{Fnd.} & \textbf{Miss.} \\
\midrule
\textit{Orchard Keeper} (1965) & 22 & 10 & 12 & 20 & 2 \\
\textit{Outer Dark} (1968) & 32 & 18 & 14 & 24 & 8 \\
\textit{Child of God} (1973) & 22 & 12 & 10 & 20 & 2 \\
\textit{Suttree} (1979) & 67 & 16 & 51 & 59 & 8 \\
\textit{Blood Meridian} (1985) & 62 & 27 & 35 & 49 & 13 \\
\textit{Pretty Horses} (1992) & 30 & 4 & 26 & 27 & 3 \\
\textit{The Crossing} (1994) & 38 & 10 & 28 & 33 & 5 \\
\textit{Cities/Plain} (1998) & 19 & 4 & 15 & 15 & 4 \\
\textit{No Country} (2005) & 11 & 6 & 5 & 6 & 5 \\
\textit{The Road} (2006) & 26 & 17 & 9 & 20 & 6 \\
\textit{The Passenger} (2022) & 19 & 0 & 19 & 19 & 0 \\
\textit{Stella Maris} (2022) & 16 & 0 & 16 & 16 & 0 \\
\midrule
\textbf{Total (12 novels)} & \textbf{364} & \textbf{124} & \textbf{240} & \textbf{308} & \textbf{56} \\
\bottomrule
\end{tabular}
\caption{Textual biblical findings per novel (including 15 signposted references; see text).  Att.\ = attested in published scholarship; Un.\ = unattested; Fnd.\ = independently discovered by the pipeline; Miss.\ = attested but not independently discovered.  An additional 253 non-textual allusions (structural, sacramental, character-level) are documented in the supplementary workbook.}
\label{tab:summary}
\end{table}

\paragraph{Anatomy of the misses}
To understand where the pipeline's 54\% recall ceiling comes from, we audited every missed allusion to determine whether the relevant McCarthy passage had ever been shown to an LLM reviewer alongside a KJV passage.  Of the missed allusions in the supplementary catalogue, approximately 94\% were \emph{retrieval failures}: the IDF and phrase-matching stages never proposed the McCarthy chunk for pairing with any KJV chunk, so no LLM ever saw the passage.  Two examples illustrate the pattern.  In \textit{Suttree}, McCarthy writes ``all this detritus slid from the city on the hill''---an unmistakable echo of Matthew~5:14 (``A city that is set on an hill cannot be hid'') and a parodic inversion in its association with modern American exceptionalism---but \emph{city} and \emph{hill} both fall below the IDF rarity threshold, and the KJV's \emph{set} is not included in the stop-word list used for n-gram matching, breaking the phrase-level overlap that would otherwise trigger a pairing.  In \textit{The Road}, the boy's reply ``I am the one'' echoes the divine self-declaration of Exodus~3:14 (``I am that I am''), yet the phrase is built entirely from the most common words in English and cannot generate an IDF signal.  The remaining ${\sim}$6\% did reach an LLM, but in every case the McCarthy chunk had been paired with the \emph{wrong} KJV chunk---a passage that happened to share vocabulary but did not contain the expected source verse.  In two of these cases the LLM nonetheless rated the mismatched pairing \textbf{Moderate}, detecting a thematic overlap even through the wrong lens.  In no case in the available data did an LLM receive the correct McCarthy--KJV pairing and reject it.  The implication is striking: when given the right passages to compare, the LLM almost never misses.  The pipeline's 54\% recall ceiling is set not by the quality of the LLM's literary judgement but by the coverage of the mechanical retrieval stages that decide which passage pairs the LLM is shown for side-by-side inspection.

\bigskip
\subsection{Per-Novel Findings}
\label{sec:per-novel}

Table~\ref{tab:summary} shows that the distribution of allusions across the corpus is uneven in revealing ways.  The Appalachian novels (1965--1979) yield 143 of the 364 textual findings.  \textit{The Orchard Keeper} has a comparably high attested rate thanks to \citeauthor{cowart-2021-allusive}'s (\citeyear{cowart-2021-allusive}) detailed allusive reading, while \textit{Suttree} alone contributes 51 findings---76\% of them unattested, despite the novel's extensive critical history.  \textit{Blood Meridian} draws on the widest range of biblical books and has the largest number of attested-but-missed allusions (13), reflecting both the novel's density and the depth of its scholarship; the embedding track recovers previously uncatalogued verbal echoes such as ``the way of the transgressor is hard'' (Proverbs~13:15) and ``horsemen riding upon horses, all of them desirable young men'' (Ezekiel~23:12; see \S\ref{sec:highlights}), while Gemini adds readings that require narrative rather than lexical overlap, notably the allusion to Moses striking the rock in the novel's enigmatic epilogue \citep{daugherty-1992-gravers, bloom-2000-how}.  The Border Trilogy marks a quantifiable shift in register (see \S\ref{sec:patterns}): direct quotation and parodic homage largely disappear, and the dominant modes become thematic transplantation and verbal echo.  \textit{The Road} has the highest proportion of attested allusions (65\%), reflecting the intensity of its critical reception.  \textit{The Passenger} and \textit{Stella Maris}---McCarthy's final novels, with no biblical-critical scholarship to date---yield 35 textual allusions between them, none for which scholarly attestation could be identified, including direct quotations (``The wicked flee when none pursue,'' ``In the beginning was the word'') and the parodic substitution discussed in \S\ref{sec:highlights}.

\bigskip
\subsection{Six Unattested Findings}
\label{sec:highlights}

To make the aggregate numbers concrete, we examine in detail six unattested findings drawn from five decades of McCarthy's career.  Our examples are chosen to illustrate the range of allusive techniques the pipeline can recover, as well as the difficulty of confirming or dismissing some types of possible allusion even with careful human review.  A number of the unattested allusions are verbatim or near-verbation borrowings, incorporated without signposting into narration and dialogue, as in examples 1--3 below.  Owing to the prominence of the Bible in Western culture, these are difficult to dismiss as mere coincidence or confounding by way of a third source.  Other of the claimed allusions show subtle register collision rather than striking phrase matches, and thus leave room for debate as to the intentionality.  Here the importance of the thematic echos marked by the pipeline might well be weighed differently by other readers.  The examples 4--6 are of this latter type.

\paragraph{1. \textit{Blood Meridian} / Ezekiel 23:6, 12}
The Glanton gang ride into a Mexican village: ``horsemen riding upon horses, all of them desirable young men.'' The phrase reproduces Ezekiel~23:12 verbatim (``horsemen riding upon horses, all of them desirable young men'').  Ezekiel~23 is a denunciation of idolatry figured as harlotry, which casts the gang's predations under a prophetic moral judgement the surrounding text otherwise withholds. The chapter is sufficiently obscure that this connection is unlikely to occur to even an attentive reader without concordance work.

\paragraph{2. \textit{The Crossing} / Matthew 7:14}
Late in the novel Billy Parham passes five horsemen in the dark: ``As if the closeness of the dark and the \emph{straitness of the way} had made of them confederates.'' The phrase is a near-verbatim borrowing from Matthew 7:14 (``strait is the gate, and \emph{narrow is the way}''), but with a characteristically McCarthian inversion: in the Gospel, the strait way is the path of salvation found only by the few; in \textit{The Crossing} it is what makes Billy and the unknown horsemen \emph{confederates} in darkness.  Interestingly, McCarthy also held at least one biographical writing on Andr\'e Gide in his personal library \citep{crews-2017-books}, who famously also borrowed from this passage in Matthew for his novel, \textit{Strait is the Gate } (or, \textit{La Porte \'Etroite} / ``\'etroite est la porte, resserr\'e le chemin'').  

\paragraph{3. \textit{Stella Maris} / 1 Corinthians 13:13}
In dialogue between Alicia and her psychiatrist: ``Language, art, mathematics, \ldots\ / And the greatest of these I take it is mathematics. / Well. I’m a mathematician.''  This is a distinct reworking of Paul's ``And now abideth faith, hope, charity, these three; but the greatest of these is charity.'' (1~Corinthians~13:13), transposing the Pauline triad onto Alicia's, and crowning mathematics in the place Paul gives to charity.

\paragraph{4. \textit{Suttree} / Genesis 22:6--12}
In a midnight dream-vision Suttree is halted in an alley: ``I was stopped by a man I took to be my father $\ldots$ I would go by but he has stayed me with his hand.  $\ldots$ The knife he held severed the pallid lamplight like a thin blue fish  $\ldots$ Yet it was not my father but my son who accosted me with such rancorless intent.'' The scene operates through inversion of the Akedah, the only biblical tableau in which a knife, a hand, and a father-son pair converge in a single moment of arrested sacrifice.  McCarthy reproduces the constitutive elements but redistributes them: in Genesis 22 Abraham's hand is stayed by an angel ("Lay not thine hand upon the lad") while here the narrator is stayed by the knife-bearing figure, and the final clause the familial roles are revealed inverted. The affectless quality Suttree attributes to the approach ("rancorless intent") preserves the register of dutiful obedience of Abraham's bearing in the KJV narrative.

\paragraph{5. \textit{Child of God} / Numbers 16:30--32}
The discovery of Lester Ballard's gruesome collection in the cave at the conclusion of the novel---``He looked in time to see his span of mules disappear into the earth taking the plow with them. He crawled with caution to the place where the ground had swallowed them \ldots\''--- is identified by our pipeline as a transposition of Korah's punishment in Numbers 16:30--32 (``the earth opened her mouth, and swallowed them up''). The matched textual echo of the ground ``swallowing'' one's earthly possessions could well be dismissed as generic subduction imagery, of which there are many examples other even in other books of the Bible.  Yet the theological parallel in terms of the relevation and judgement of a community's shared guilt---though by way of collective indifference in \textit{Child of God} compared to rebellion in Numbers---builds a reasonable thematic link.  That ultimately the mules are never found, despite the sheriff's operation to recover the human bodies, builds the mystical tone. 

\paragraph{6. \textit{All the Pretty Horses} / Jeremiah 31:33}
After his escape from captivity John Grady dreams that the horses come upon ``an antique site where some ordering of the world had failed and if anything had been written on the stones the weathers had taken it away again \ldots\ Finally what he saw in his dream was that the order in the horse's heart was more durable for it was written in a place where no rain could erase it.''  The lexical surface is thin, but the structural embedding is unmistakably Jeremiah~31:31--33 :``which my covenant they brake $\ldots$ I will put my law in their inward parts, and write it in their hearts''.  The image of writing erased from stones reflects the Mosaic tablets whose covenant was broken by the fathers of the Israelites, after which we have a characteristic McCarthy transposition that puts horses in the place of the tribes.  \citet{mundik-2017-bloody} reads the same passage through Gnostic emanationism but does not connect it to the new-covenant tradition.

\bigskip
\subsection{Patterns Across the Corpus}
\label{sec:patterns}

A reproducible procedure for allusion detection makes it possible to ask aggregate questions about an author's biblical engagement that no individual close reading can answer.  The catalogue contains 612 individual book references, since some allusions cite multiple verses.  Five books account for nearly half: Matthew (16\%), Genesis (13\%), Revelation (10\%), Luke (6\%), and John (6\%).  The Old and New Testaments are roughly equally represented overall (49\% OT, 51\% NT), and the balance holds steady across all four career periods, varying by no more than ten percentage points.  McCarthy is not, then, an author who writes ``out of'' the Old Testament or the New, but one who draws on both halves of the canon in roughly equal measure.  Within this balance, the Synoptic Gospels account for 24\% of references, the Pentateuch for 23\%, and the wisdom literature plus Revelation for 24\%---a combination that maps onto McCarthy's recurring preoccupations with origins, parables, theodicy, and apocalypse.

The connection-type breakdown shows a clear shift in McCarthy's allusive technique across his career (Table~\ref{tab:type-by-period}).  The Appalachian novels deploy the full range of textual modes, including 14 direct quotations and 22 parodic homages.  By the Border Trilogy, direct quotation has nearly disappeared (2 instances) and parodic homage has dropped to 3; the dominant modes are thematic transplantation and verbal echo.  The Late period continues this trend.  Critics have observed the shift qualitatively---McCarthy's biblical engagement becomes more diffuse and less explicit as his style matures \citep{cant-2008-mccarthy, lincoln-2009-mccarthy}---but it has not previously been quantified across the full corpus.  The decline of direct quotation after \textit{Suttree} is particularly striking: the early novels stage characters \emph{quoting} scripture aloud, while the later novels rework biblical material at the level of cadence and image without explicit signposting.  Individual novels show distinctive single-book signatures as well: \textit{Blood Meridian} draws disproportionately on Revelation, and the Border Trilogy on Genesis (Eden, Sodom, Abraham), while \textit{The Road}'s thematic engagement with Job---well established in the criticism---operates more through theological argument than through verse-level allusion, and \textit{Suttree} is distinguished less by any single-book concentration than by the sheer breadth of its scriptural range.

\begin{table}[t]
\centering
\small
\begin{tabular}{lrrrr}
\toprule
\textbf{Connection type} & \textbf{App} & \textbf{BM} & \textbf{Bor} & \textbf{Late} \\
\midrule
Thematic transplantation & 28 & 17 & 42 & 29 \\
Verbal echo & 33 & 17 & 33 & 22 \\
Inverted allusion & 22 & 15 & 5 & 13 \\
Parodic homage & 22 & 4 & 3 & 1 \\
Transformed imagery & 17 & 5 & 1 & 2 \\
Direct quotation & 14 & 1 & 2 & 1 \\
Register collision & 8 & 5 & 4 & 6 \\
\bottomrule
\end{tabular}
\caption{Textual connection types by period.  App = Appalachian (4 novels); BM = \textit{Blood Meridian}; Bor = Border Trilogy (3 novels); Late = Late period (4 novels).}
\label{tab:type-by-period}
\end{table}

\bigskip
\section{Discussion}
\label{sec:discussion}

The McCarthy--KJV catalogue constructed here is, by design, less a final account of one author's biblical inheritance than a worked calibration of the instrument by which such accounts might be conducted at scale.  What makes the calibration worth reporting at this length is not the novelty of any individual finding---though we believe several are of independent literary interest---but what the overall pattern of successes and failures reveals about the feasibility of a larger enterprise: the AI-assisted, statistically grounded study of literary influence noted in the Introduction.  The design of that enterprise requires precisely the type of insights on the efficacy of LLMs within stylometric pipelines for allusion detection that the present study delivers.

\paragraph{From one author to a population}
McCarthy is a natural first case study precisely because his biblical engagement is so richly documented in the critical literature, furnishing a ground truth against which pipeline performance can be measured with some confidence.  But there is no reason to suppose that the pipeline's capabilities are specific to McCarthy's style or to the KJV as a source text, and the ambition that motivates the present work extends well beyond a single author--source pairing.  The prospect before us is one of drawing insights from study at scales beyond the reach of individual close reading---not as a rival to the foundational work of scholars such as \citet{frye-1982-great-code} and \citet{alter-2010-pen}, but as a complement to it, supplying the comparative and aggregate evidence that close reading by its nature cannot.  How deeply has the KJV shaped American fiction as a whole?  How does that influence vary across regions, genres, and decades? And which biblical books have proved most productive for which literary traditions?

Epidemiology offers a useful analogy here: a template for the type of knowledge such a programme would produce.  Epidemiological reasoning proceeds not by tracing the precise causal pathway by which one individual contracted a disease but by studying differences in health outcomes across contrasting populations, and from those contrasts identifying the genetic, environmental, and behavioural factors that dispose certain groups to higher risk \citep{broadbent-2013-philosophy}.  The knowledge it generates is irreducibly population-level: it characterises the distribution of risk and the factors that condition it, not the particular natural history of any single case---it brings forth insights into those things that are common to us all and those that vary across the stratifications that divide us.  A programme of AI-assisted literary influence study would produce knowledge of the same philosophical type \citep{broadbent-2013-philosophy}, mapping the distribution of biblical engagement across a corpus of authors, and from contrasts between regions, periods, genres, and traditions, identifying the social and historical factors that dispose one body of writing to draw more heavily on scripture than another.  The pipeline, in this analogy, is the survey instrument; the McCarthy catalogue is its field trial.

\paragraph{Calibrating the instrument}
Any such programme would begin, as epidemiological studies must, from a substantive hypothesis---for example, that the direct influence of the KJV on American literary fiction has waned over the twentieth century in step with declining church attendance and religious schooling---together with a target corpus and supporting data against which that hypothesis can be evaluated.  The statistical framework appropriate to the inquiry is a Bayesian model with two coupled components.  The first is a \emph{substantive model} encoding the hypothesis itself: a multilevel regression \citep{greenland-2000-multilevel}, say, with time period, region, genre, and the author's relevant biographical details as covariates, and with causal and confounding pathways distinguished by prior structure.  The second is an explicit \emph{measurement model} for the pipeline's sensitivity and specificity, parameterised by allusion type, genre, period, and any other strata for which calibration data are available \citep{joseph-1995-bayesian}.  It is the measurement model that makes the enterprise viable despite the pipeline's imperfect recall: raw allusion counts need not be taken at face value but can be adjusted through estimated detection probabilities, so that an author whose biblical engagement runs predominantly through structural typology---a category the pipeline detects poorly---is not systematically under-counted.

Our results supply workable first estimates for the parameters required for \emph{a priori} experimental design of such a modelling study.  The stratified figures in Table~\ref{tab:by-type}---80\% recall for register collisions, declining to 30\% for transformed imagery---give initial point estimates of sensitivity by allusion type, and the audit of missed allusions reported in \S\ref{sec:results} localises the source of these losses to the mechanical retrieval stages rather than to LLM judgement: in no audited case did an LLM receive the correct passage pair and reject it.  The recall ceiling is set not by the quality of the literary reasoning applied downstream but by the coverage of the mechanical retrieval stages that determine which passage pairs the LLM is shown.  This is at once the pipeline's most significant limitation and its most encouraging finding: retrieval coverage is straightforwardly improvable---and, in the limit, dispensable---whereas unreliable literary judgement on the part of the LLM would pose a deeper problem altogether, one less amenable to engineering.

Learning of both model components would proceed jointly.  Human review of pipeline findings, historical research into author notebooks and libraries, and LLM-directed literature search constrain the measurement-model parameters---how often does the pipeline miss a real allusion, and how often does it fabricate one?---while the substantive model propagates that uncertainty forward into the quantities of interest.  The two components cannot usefully be separated: the answer to the substantive question depends on how much we trust the instrument, and the most informative places to calibrate the instrument depend on where the substantive question is most sensitive to measurement error.

\paragraph{Adaptive design}
The joint model just described has four principal levers by which its parameters can be tightened, each with a different cost profile.  \emph{Human review} of individual pipeline findings and \emph{historical research} are the most expensive per unit but yield the highest-quality ground truths.  \emph{LLM-directed literature search} across scholarly databases is cheaper, and can establish or rule out attestation status for large batches of findings, tightening the measurement model without human labour.  \emph{Full chunk-by-chunk LLM scanning}---bypassing the mechanical retrieval filter altogether and presenting every novel-chunk / source-chunk pair directly to an LLM reviewer---is the most informative per novel, since it removes the retrieval bottleneck just described, but at a cost that is moderate per work and prohibitive across thousands; for a single novel--source pairing the number of chunk-pairs is in the low thousands, well within the throughput of current models, yet at population scale the combinatorial explosion makes a pre-filter essential.

Within a given budget of API calls and human review time, Bayesian adaptive design \citep{chaloner-1995-bayesian-design, cui-2019-adaptive} can draw efficiently upon all three levers, choosing at each step not only which lever to pull but which work from the corpus to apply it to.  The criterion is expected information gain: if the substantive model is most uncertain about a particular genre or period, the next work is drawn from that stratum; if the measurement model is weakest for a particular allusion type, human review is directed to findings of that type; if a work sits at the boundary between two competing hypotheses, full LLM scanning is deployed to maximise the evidential yield.  Effective adaptive strategies may thereby allow the programme to reach its learning goals from a calibration sample that is a but small fraction of the target corpus.

\paragraph{Rational and statistical inquisition of the LLM components}
The statistical framework just described treats LLM judgement as a measurement instrument whose error properties can be estimated and propagated forward into quantities of interest.  A distinctive epistemic question remains, however: the LLMs at the centre of the pipeline have themselves absorbed much of the literary commentary against which we hope to evaluate them.  When an LLM identifies a biblical allusion in McCarthy, we cannot in general distinguish whether the model is reading both texts and arriving at the connection by inference, recalling a scholar's prior identification from its training data, or drawing on a connection so culturally pervasive as to require neither.  Our \textbf{Attested}/\textbf{Unattested} tagging bounds this ambiguity but does not resolve it.

To probe the mechanism directly we conducted a contamination experiment on three frontier models (Gemini 3.x Pro, GPT-5, Claude Sonnet 4.5), reported in full in Appendix~\ref{sec:appendix-contamination}.  Thirteen test items spanned four categories: \emph{attested} McCarthy passages with documented allusions; \emph{unattested} passages where our pipeline had surfaced echoes no scholar has catalogued; \emph{synthetic} McCarthy-pastiche passages we authored via a different LLM minutes before the test, with KJV echoes deliberately planted; and \emph{null} pastiche with no biblical content.  The synthetic items are the diagnostic case: by construction they cannot exist in any training corpus, so any detection is necessarily cross-textual analysis rather than scholarly recall.  All three models scored 12/12 on the synthetic items, in several cases naming multiple parallel verses and flagging the specific grammatical transpositions we had planted; false-positive rates on the null items were near zero.  Strikingly, the one item all three models missed was the most heavily documented reading in the set---\textit{The Road}'s ``the clocks stopped at 1:17'' / Revelation~1:17---which hinges on a numerical coincidence rather than lexical overlap.  A scholarship-recall mechanism would have scored attested items highest and synthetic items lowest; we observe the opposite.

This experiment exemplifies what we term \emph{rational inquisition} of the model: asking whether its observed pattern of successes and failures is consistent with the mechanism we suppose it to be using.  The complementary \emph{statistical inquisition}---asking whether the data-generating process is plausibly consistent with the technical assumptions of our model---would involve running the pipeline on works published before any candidate source could plausibly have influenced them, a \emph{negative-control} design \citep{lipsitch-2010-negative}, and cross-checking against authors whose specific influences are documented in their own notebooks or commonplace books \citep{lowes-1927-road, crews-2017-books}.  Together these three probes---synthetic pastiche, negative controls, and ground-truth notebooks---would constitute a three-way calibration quantifying both directions of error.  That we cannot yet fully resolve how the models are ``reading'' need not prevent us from using them productively under calibrated oversight of their performance.  What matters for the wider programme is not that the instrument's internal workings be fully transparent but that its external behaviour be predictable enough to model statistically---and the contamination probe suggests that, within the domain tested, it is.

\paragraph{Pipeline improvements and infrastructure}
Since the missed-allusion audit identifies mechanical retrieval rather than LLM judgement as the binding constraint on recall, the most productive near-term improvements target the retrieval stage.  A single-rare-word anchor pass would let high-IDF vocabulary trigger candidate pairs without requiring multi-word overlap; top-$K$ KJV chunk presentation would catch echoes when the best n-gram match is not the most apt pairing; and lemmatisation, an onomastic index, and a parallel pass against the Reina-Valera and New Jerusalem Bibles would close further gaps.  None of these would alter the pipeline's architecture or its LLM review stages, which on the evidence available are performing their proxy-reviewer role with high fidelity.  Scaling to a population of authors, however, raises a different order of difficulty.  Online repositories cover only a thin and selection-biased slice of twentieth-century fiction; restricting analysis to these databases for sheer expediency risks both statistical bias through confounding by selection and cultural bias through omission.  Comprehensive coverage of the kind the wider programme demands can only be assembled through national libraries and copyright-holding publishers, most plausibly via federated-access models of the sort developed in public health for sensitive medical records.

\paragraph{Scope of the present contribution}
We have taken one author and one source text and produced a calibrated, provenance-tagged catalogue that surfaces 349 allusions and reveals corpus-level patterns---the steady OT/NT balance, the decline of direct quotation after \textit{Suttree}, the per-novel biblical signatures---that would be difficult to establish by other means.  But the catalogue's more important function is as a worked example of what it would take to conduct such an analysis properly at scale.  The dual-track architecture, the stratified evaluation, the contamination probe, and the adaptive-design framework are all transferable to other author--source pairs; the numbers in our tables should be read in that spirit.  They are not a final account of biblical influence in McCarthy but a first calibration point for a quantitative study of literary influence that does not yet exist---but that, as we hope to have shown, the tools and methods now at hand make achievable.

\bigskip
\section*{Limitations}
\label{sec:limitations}

\paragraph{Source text and language restrictions}
We use only the King James Bible as source text.  McCarthy's biblical language may also reflect other translations---his ``witless paraclete'' in \textit{Outer Dark} suggests the New Jerusalem Bible and his Catholic upbrining the Douay-Rheims---and allusions to non-KJV phrasings will be missed.  Separately, McCarthy's border novels contain extensive untranslated Spanish; the pipeline filters Spanish tokens to prevent false matches, which means that allusions operating through Spanish-language phrasing, including echoes of the Reina-Valera Bible, are systematically excluded.

\paragraph{Allusion types the pipeline cannot catch}
The embedding track requires shared rare vocabulary as its entry point and consequently cannot routinely surface allusions that operate through common words, structural parallels, or character typology.  The register scan and Gemini long-context track partially address this blind spot, but some categories remain inherently difficult: a character functioning as a Moses or Job figure without distinctive shared vocabulary will not be detected by any track if the parallel is subtle.

\paragraph{LLM pre-training and the Unattested label}
The LLM components have absorbed extensive literary commentary during pre-training.  Findings classified as \textbf{Unattested} may therefore reflect allusions that are well known to scholars but absent from the indexed databases and published accessible scholarship.  We mitigate this risk by cross-checking against multiple sources including the Semantic Scholar API and a hand-compiled reference table drawing on over thirty publications---monographs \citep{cant-2008-mccarthy, lincoln-2009-mccarthy, mundik-2017-bloody, potts-2015-sacrament, broncano-2014-religion, crews-2017-books}, edited volumes \citep{noble-2020-bible}, journal articles and book chapters spanning five decades from \citet{schafer1977hard} to \citet{cowart-2021-allusive}---but acknowledge that an exhaustive bibliographic audit would require access to several further works that we were unable to obtain.  The mechanistic question---whether the LLMs are performing cross-textual analysis or recalling memorised commentary---is addressed by the contamination probe reported in \S\ref{sec:discussion} and Appendix~\ref{sec:appendix-contamination}, whose findings favour the former interpretation but rest on a small item set and would benefit from replication at larger scale.

\paragraph{Subjectivity, reproducibility, and generalisability}
LLM literary judgement introduces a form of subjectivity that differs from but does not eliminate the subjectivity of human close reading.  We do not measure inter-annotator agreement between models, and the pipeline's output will vary with model versions---a limitation inherent to any methodology that relies on rapidly evolving foundation models.  The pipeline has been developed and tested on a single author--source pair (McCarthy--KJV); its effectiveness on other authors, literary traditions, or languages is untested, and the IDF thresholds, embedding cutoffs, and prompt designs would likely require recalibration for other applications.

\bigskip

\section*{Ethics Statement}
\label{sec:ethics}

\paragraph{Texts and copyright}
McCarthy's novels remain under copyright; we use them under fair-use provisions for non-commercial scholarly research.  Quoted excerpts are short, transformative, and presented for literary criticism.

\paragraph{Computational cost}
The full pipeline costs approximately \$50--80 in API calls and runs in 8--12 hours on consumer hardware---deliberately kept cheap to make this style of analysis accessible outside well-resourced research groups.

\bibliographystyle{compling}
\bibliography{references}

\appendix

\bigskip
\section{Pipeline Details}
\label{sec:appendix-pipeline}

This appendix documents the pipeline architecture in sufficient detail for replication, covering model specifications, hyperparameters, and representative prompt templates.  All prompts use temperature~0 to maximise reproducibility.

\bigskip
\subsection{Corpus Preparation and IDF}
\label{sec:app-idf}

IDF scores are computed across a reference corpus of approximately 800 texts spanning classical, early modern, and twentieth-century literature. A word or phrase qualifies as \emph{rare} if its IDF exceeds 4.0 (i.e., it appears in $\leq$15 texts). McCarthy-side floors are set lower (word IDF $\geq$1.5, phrase IDF $\geq$0.0) to capture McCarthy's use of common biblical vocabulary in distinctive combinations. Before inclusion in the matching database, candidates are filtered to remove OCR artefacts, Spanish-language tokens, and proper names.

\bigskip
\subsection{Embedding and Matching}
\label{sec:app-embed}

Each rare word or phrase occurrence is recorded with a context snippet ($\pm$4 words, truncated at clause- or sentence-breaking punctuation but extending through commas) and embedded using two Voyage AI models:

\begin{itemize}
    \item \texttt{voyage-4}: semantic embedding for sense disambiguation
    \item \texttt{voyage-4-large}: contextual embedding for broader usage similarity
\end{itemize}

\noindent Embeddings are computed in batches of 128. The resulting database contains approximately 200K word occurrences and 2.4M phrase occurrences.

Cross-corpus matching operates in four directions with the following cosine similarity thresholds:

\begin{table}[h]
\centering
\small
\begin{tabular}{lcc}
\toprule
\textbf{Direction} & \textbf{Semantic} & \textbf{Context} \\
\midrule
Word $\leftrightarrow$ Word & $\geq 0.88$ & $\geq 0.40$ \\
Phrase $\leftrightarrow$ Phrase & $\geq 0.88$ & $\geq 0.40$ \\
Word $\leftrightarrow$ Phrase & $\geq 0.82$ & $\geq 0.40$ \\
Phrase $\leftrightarrow$ Word & $\geq 0.82$ & $\geq 0.40$ \\
\bottomrule
\end{tabular}
\caption{Cosine similarity thresholds for cross-corpus matching.}
\label{tab:thresholds}
\end{table}

\bigskip
\subsection{Density Mapping}
\label{sec:app-density}

Sliding windows at three scales scan both texts simultaneously:

\begin{itemize}
    \item \textbf{Narrow} (500 characters): minimum 2 distinct matches, step = 125 chars
    \item \textbf{Medium} (2{,}000 characters): minimum 4 distinct matches, step = 500 chars
    \item \textbf{Large} (5{,}000 characters): minimum 6 distinct matches, step = 1{,}250 chars
\end{itemize}

\noindent Clusters detected at multiple scales receive cross-scale boosting; a quality floor of 0.5 filters low-confidence clusters.

\bigskip
\subsection{LLM Models and Parameters}
\label{sec:app-models}

\begin{table}[h]
\centering
\small
\begin{tabular}{llrr}
\toprule
\textbf{Stage} & \textbf{Model} & \textbf{Tokens} & \textbf{Workers} \\
\midrule
Haiku screen & \texttt{claude-haiku-4-5} & 1{,}024 & 10 \\
GPT-5.4 deep & \texttt{gpt-5.4} & 2{,}048 & 5 \\
Biblical scan & \texttt{claude-sonnet-4-6} & 1{,}024 & 30 \\
Verse confirm & \texttt{claude-sonnet-4-6} & 512 & 30 \\
Biblical triage & \texttt{gpt-5.4} & 2{,}048 & 10 \\
Gemini verify & \texttt{gemini-3.1-pro} & --- & 1 \\
SI compilation & \texttt{gpt-5.4} & 12{,}000 & 1 \\
\bottomrule
\end{tabular}
\caption{Model specifications per pipeline stage. All stages use temperature~0. Token limits are maximum completion tokens.}
\label{tab:models}
\end{table}

\bigskip
\subsection{Prompt Templates}
\label{sec:app-prompts}

We provide representative prompts for the four core LLM stages.  Minor formatting details are omitted for brevity; full prompts are available in the code repository.

\paragraph{Haiku unprimed screen (Stage~5)}
The system prompt instructs the model to act as a scholar of comparative literature and emphasises the critical distinction between thematic similarity and genuine intertextuality:

\begin{quote}
\small
\textit{``Two passages about SIMILAR THEMES (death, winter, cities, rivers, decay) is NOT evidence of intertextuality. These are universal literary subjects. Evidence requires SPECIFIC SHARED PHRASING---sentences or clauses with matching syntactic structure, unusual vocabulary appearing in both, or distinctive images that are clearly reworked rather than independently invented.''}
\end{quote}

\noindent The model receives both passages without being told which words triggered the match, and rates the connection as \textbf{Strong}, \textbf{Moderate}, \textbf{Weak}, or \textbf{None}, with required quotation of specific phrases from both texts.

\paragraph{GPT-5.4 deep analysis (Stage~6)}
The system prompt extends the taxonomy of connection types to six categories and instructs the model to treat convergent evidence as stronger than any single signal:

\begin{quote}
\small
\textit{``When MULTIPLE signals co-occur---e.g.\ a transplanted character AND a decomposed coinage AND a matching seasonal setting---treat the combination as stronger evidence than any single signal. McCarthy's allusions often work through transformation rather than quotation.''}
\end{quote}

\noindent The model receives line-numbered passages (prefixed M for McCarthy, A/K for source) and returns structured JSON with exact quotations from both texts, line references, and explanatory notes for each finding.

\paragraph{Biblical register scan (Stage~7)}
The system prompt instructs the model to read McCarthy passages \emph{without any KJV text} and flag biblical language across six categories: direct quotation, KJV cadence, biblical vocabulary, prophetic register, liturgical patterns, and transformed allusions. For each flagged passage, the model identifies the likely KJV source at book-and-verse specificity where possible:

\begin{quote}
\small
\textit{``Look for: DIRECT QUOTATION or near-quotation of KJV verses (even partial); KJV CADENCE: archaic syntax patterns characteristic of the KJV (e.g.\ `and he X, and Y, and Z' parataxis; `sufficient unto the day'; `the X thereof'); BIBLICAL VOCABULARY: words rare in modern English but common in KJV (e.g.\ `straitness', `raiment', `smote', `begat', `dominion', `iniquity').''}
\end{quote}

\paragraph{Gemini long-context verification (Stage~10)}
Each McCarthy novel is loaded in full alongside approximately half the KJV (${\sim}$500K--650K tokens per call). The prompt requests identification of \emph{all} passages that ``echo, quote, transform, invert, or allude to specific KJV Bible passages,'' with required exact quotation from both texts and verse-level attribution. Four complementary passes cover: (a)~Genesis--Job, (b)~Psalms--Revelation, (c)~hint-guided review using the pipeline's matched rare words, and (d)~a review of existing pipeline findings for false positives and omissions.

\bigskip
\subsection{Quote Verification and Deduplication}
\label{sec:app-dedup}

All KJV quotations attributed to findings are verified by searching for 4-word fragments from each quote in the full KJV text.  Deduplication then proceeds at two levels.  First, multiple chunk pairings that target the same McCarthy passage are collapsed by fuzzy word-overlap matching ($\geq$3 shared distinctive words at $\geq$30\% overlap of the shorter quote).  Second, multiple GPT-5.4 classifications of the same passage (e.g., as both ``verbal echo'' and ``register collision'') are collapsed, retaining the version with the most detailed analysis.

\bigskip
\subsection{Bibliographic Verification}
\label{sec:app-biblio}

Each allusion is checked against the Semantic Scholar API \citep{kinney-etal-2023-semantic-scholar} using queries combining the novel name, allusion terms, and biblical book name (rate-limited to 1 request per second).  Results are cross-referenced against the specialist-compiled reference table of 126 entries.  GPT-5.4 then assigns each finding to \textbf{Attested} or \textbf{Unattested}, with strict provenance separation: findings independently discovered by the pipeline are tagged with their discovery source (Pipeline, Gemini, or both), while reference-table entries not independently recovered are tagged as Missed.  GPT-5.4 is instructed to prefer ``No citation found'' over fabrication and to mark uncertain citations with \texttt{[UNVERIFIED]} flags.  A second round of Semantic Scholar queries verifies citations provided by GPT-5.4.

\bigskip
\section{Training-Data Contamination Probe}
\label{sec:appendix-contamination}

This appendix documents the contamination experiment summarised in \S\ref{sec:discussion}.  The full test harness, items, prompts, and raw model outputs are available in the code repository.

\bigskip
\subsection{Question and design}
\label{sec:app-cont-design}

The experiment is designed to discriminate between two possible mechanisms by which an LLM in our pipeline might attribute a biblical allusion to a McCarthy passage: (a) genuine cross-textual analysis between the McCarthy passage and the model's internal KJV representation, and (b) memorised retrieval of scholarly commentary the model absorbed during pre-training. We tested three frontier models---\texttt{gemini-3.1-pro-preview}, \texttt{gpt-5}, and \texttt{claude-sonnet-4-5}---on thirteen short test passages drawn from four categories.

Each item was presented to each model under the same neutral prompt, which did not mention McCarthy, the King James Bible, or any other prime. The model was asked to identify any literary or biblical allusions, quotations, or intertextual echoes in the passage, and was instructed to prefer ``uncertain'' or ``no allusion'' over fabrication. Temperature was 0; each item was run once per model.

\paragraph{Type A — Attested (3 items)}
Real McCarthy passages with biblical allusions documented in the McCarthy scholarship surveyed for our reference table. Both memorisation and genuine analysis should succeed here; failure indicates a basic detection problem. The items are: \textit{The Road} opening (``the clocks stopped at 1:17'') $\rightarrow$ Rev~1:17; \textit{Blood Meridian} dust-of-the-earth speech $\rightarrow$ Gen~2:7 and Mark~4:10--13; \textit{Outer Dark} title description $\rightarrow$ Matt~8:12 / 22:13 / 25:30.

\paragraph{Type B — Unattested (3 items)}
Real McCarthy passages where our pipeline surfaced biblical echoes that no scholar in the sources we surveyed has documented. Memorisation of McCarthy scholarship cannot succeed because there is nothing scholarly to recall. The items are: \textit{The Orchard Keeper}'s ``no olive branch but hard coin'' $\rightarrow$ Gen~8:11; \textit{All the Pretty Horses} horse-heart dream $\rightarrow$ Jer~31:33 and 2~Cor~3:3; \textit{Suttree} ``He give out the loaves and the fishes, he howled. Therefore ast not what shall I put on.'' $\rightarrow$ Matt~14:19 and Matt~6:25.

\paragraph{Type C — Synthetic (4 items)}
McCarthy-pastiche passages written by a different LLM minutes before the test, each containing a deliberately planted KJV echo. These passages cannot exist in any training corpus, so any detection is necessarily textual analysis. The planted sources are: Isa~40:6 / 1~Pet~1:24 / Ps~103:15--16 (grass / flesh / flower of the field); Matt~10:29--30 (sparrow falling, hairs of the head numbered); Eccl~1:7, 1:9 (rivers running into the sea, nothing new under the sun); Job~14:1--2 (``of few days and full of trouble,'' ``cometh forth like a flower''). Each pastiche reproduces McCarthy's stylistic register (polysyndeton, elliptical dialogue, no quotation marks, archaic diction) but the planted phrases are identifiable only by reading the text against an internal KJV representation.

\paragraph{Type D — Null (3 items)}
McCarthy-pastiche prose written with deliberately no biblical echo: a pastoral river crossing, a truck on a gravel road, and a domestic interior. McCarthy-style register alone should not be enough to provoke an allusion claim; a model that hallucinates allusions on null items is doing pattern-matching on style rather than analysis on content.

\bigskip
\subsection{Results}
\label{sec:app-cont-results}

Table~\ref{tab:contamination} reports per-item, per-model outcomes. A \emph{hit} means the model named the intended source; a \emph{hedged hit} means the source appears in the response but framed as uncertain; a \emph{miss} means no mention of the correct source; a false positive on a null item means an unhedged allusion claim.

\begin{table*}[t]
\centering
\small
\begin{tabular}{llllll}
\toprule
\textbf{Item} & \textbf{Type} & \textbf{Expected} & \textbf{Gemini 3.x Pro} & \textbf{GPT-5} & \textbf{Sonnet 4.5} \\
\midrule
A1 & Attested & Rev 1:17 & Hedged & Miss & Miss \\
A2 & Attested & Gen 2:7 + Mark 4 & Hit & Hit & Hit \\
A3 & Attested & Matt 8:12/22:13/25:30 & Hit & Hit & Hedged \\
B1 & Unattested & Gen 8:11 & Hit & Hit & Partial miss \\
B2 & Unattested & Jer 31:33 / 2 Cor 3:3 & Hit & Hit & Hedged hit \\
B3 & Unattested & Matt 14:19 / Matt 6:25  & Hit & Hit & Hit \\
C1 & Synthetic & Isa 40:6 / 1 Pet 1:24 & Hit & Hit & Hit \\
C2 & Synthetic & Matt 10:29--30 & Hit & Hit & Hit \\
C3 & Synthetic & Eccl 1:7, 1:9 & Hit & Hit & Hit \\
C4 & Synthetic & Job 14:1--2 & Hit & Hit & Hit \\
D1 & Null & --- & Correct null & Correct null & Correct null \\
D2 & Null & --- & Correct null & Soft FP (hedged) & Correct null \\
D3 & Null & --- & Correct null & Correct null & Correct null \\
\bottomrule
\end{tabular}
\caption{Per-item, per-model outcomes for the contamination probe. ``Hedged'' = source named but framed as uncertain or as ``no definite allusion.''}
\label{tab:contamination}
\end{table*}

The aggregate scores are summarised in Table~\ref{tab:contamination-summary}.

\begin{table}[h]
\centering
\small
\begin{tabular}{lcccc}
\toprule
\textbf{Model} & \textbf{Att (3)} & \textbf{Un (3)} & \textbf{Syn (4)} & \textbf{Null FP} \\
\midrule
Gemini 3.x Pro & 2.5 & 2 & \textbf{4/4} & 0 \\
GPT-5 & 2 & 2.5 & \textbf{4/4} & 1 (hedged) \\
Sonnet 4.5 & 1.5 & 1 & \textbf{4/4} & 0 \\
\bottomrule
\end{tabular}
\caption{Aggregate scores. Att = attested, Un = unattested, Syn = synthetic. Hedged hits count as 0.5; partial misses as 0. The synthetic column is the diagnostic case.}
\label{tab:contamination-summary}
\end{table}

\bigskip
\subsection{Interpretation}
\label{sec:app-cont-interp}

The pattern of hits and misses across item types bears directly on the contamination question.

\paragraph{The synthetic items rule out contamination as a sufficient explanation}
All three models score 12/12 across the synthetic passages. In two cases the model spontaneously identified the specific grammatical transposition planted in the pastiche: GPT-5 noted that the ``hairs of the head numbered'' phrasing in C2 had been transferred from humans to sparrows, and Sonnet flagged the same item as an ``ironic inversion'' of the biblical reassurance. These passages were written minutes before the test and cannot exist in any training corpus; the only possible detection mechanism is reading the pastiche against the models' internal KJV representation. Scholarship recall is ruled out by construction for this category, and all three models perform equally well or better on the synthetic items than on the attested ones---the inverse of what pure scholarship-memorisation would predict.

\paragraph{The hardest item is the one with the most scholarship}
A1 (\textit{The Road} ``the clocks stopped at 1:17'' / Rev~1:17) is the only item all three models miss or hedge on. The Rev~1:17 reading is documented in standard McCarthy scholarship on \textit{The Road} (e.g., \citealp{mundik-2017-bloody}) and discussed in numerous reviews and journal articles; if scholarship-recall were the dominant detection mechanism this item should be trivial. All three models fail it because the connection hinges on a numerical coincidence---the time ``1:17'' pointing to a verse ``1:17''---rather than any lexical overlap, and frontier LLMs do not spontaneously recognise number-as-citation. A1's failure is the strongest single piece of evidence against the contamination hypothesis: the most heavily documented attested item is missed by all three models, while every self-authored synthetic item is hit cleanly. A scholarship-driven mechanism would have produced exactly the opposite ranking.

\paragraph{Null hallucination rates are low}
Gemini and Sonnet produce zero false positives on the null pastiche (3/3 each); GPT-5 produces one near-positive on D2, explicitly hedged as ``uncertain---possible Eccl~1:4, 1:11.'' All three models correctly note the McCarthy-style register (polysyndeton, paratactic syntax, archaic diction) without treating that style as evidence of biblical content. McCarthy-style prose alone is not sufficient to prime the models to project biblical allusions; the detection responds to phrase-level and structural overlap, not to register.

\bigskip
\subsection{Model-calibration profiles}
\label{sec:app-cont-calibration}

Running three models in parallel yields a clearer picture of their respective calibration profiles. Gemini is the most confidently committal, enumerating multiple candidate sources per passage and rarely hedging; it had the highest non-null hit rate (9.5/10) with zero null false positives. GPT-5 is intermediate, committing to candidates while adding explicit ``uncertain'' labels on borderline cases; it produced the only soft false positive but also caught the most secondary intertexts (e.g., Shelley's ``Ozymandias'' as a parallel to the APH horse-heart dream). Sonnet is the most conservative, often declining to commit even after naming a candidate source: on B2 it identified both Jer~31:33 and Heb~8:10 in its response but then said ``the passage does not echo specific biblical language closely enough for me to claim a definite allusion.'' This conservatism keeps Sonnet's null false-positive rate at zero but also costs it real hits, including the olive-branch / Gen~8:11 connection in B1, which it missed because once it had committed to the Prodigal Son frame it did not enumerate parallel possibilities. The calibration profiles align naturally with the roles each model plays in the production pipeline: Haiku for fast wide-net screening, GPT-5.4 for detailed analysis, Sonnet for the conservative biblical-register scan.

\bigskip
\subsection{Limitations of the probe}
\label{sec:app-cont-limits}

\paragraph{Sample size}
13 items $\times$ 3 models = 39 observations is enough to establish the qualitative pattern but not enough to support precise statistical claims about precision/recall differences between models.  A scaled version with 30--50 items per type would be a natural extension.

\paragraph{Prompt priming}
The neutral prompt explicitly mentions ``literary or biblical allusions,'' which could bias the models toward biblical attributions. The low null false-positive rates suggest the bias is small in practice, but a fully neutral version (``any intertextual echoes'') would be a cleaner replication.

\paragraph{Model versions}
The probe used \texttt{gemini-3.1-pro-preview}, \texttt{gpt-5}, and \texttt{claude-sonnet-4-5}, which are one version increment behind the production pipeline's \texttt{gemini-3.1-pro}, \texttt{gpt-5.4}, and \texttt{claude-sonnet-4-6-20250514}. We expect the central result (12/12 on synthetic items) to be robust to version increments because newer models in this family are if anything more capable on cross-textual retrieval; re-running the probe on the production versions before final submission would be straightforward.

\paragraph{Sonnet's commitment threshold}
Sonnet named both Jer~31:33 and Heb~8:10 in its B2 response but framed the answer as ``no definite allusion.'' Whether this counts as a hit or a miss is a scoring convention; we recorded it as a hedged hit. A prompt asking ``what sources could this echo?'' rather than ``what allusions are present?'' would likely surface more of Sonnet's latent detections.

\paragraph{An indirect contamination signal}
On C1 (grass / flower of the field) Gemini's first-pass response was blocked by an internal recitation filter (\texttt{finish\_reason = RECITATION}), indicating that the model recognised it would need to quote near-verbatim canonical text in order to respond. This is not a failure mode but an indirect confirmation of detection: the model recognised the KJV source clearly enough to know its natural response would contain protected scriptural language. On the second pass with a paraphrase-only instruction, Gemini named Isa~40:6, 1~Pet~1:24, and Ps~103:15--16 correctly. GPT-5 and Sonnet had no recitation-filter issues and named all three sources on the first pass.

\bigskip
\section{Full Allusion Catalogue}
\label{sec:appendix-catalogue}

The complete supplementary information table of all 527 biblical connections identified across the twelve-novel corpus is provided as an Excel workbook, \texttt{mccarthy\_biblical\_allusions\_SI\_final.xlsx}, accompanying this submission.  This includes 364 textual findings analysed in the main paper (of which 349 are classified as allusions and 15 as signposted references), 8 duplicate entries retained for traceability but excluded from all counts, and 155 non-textual allusions (structural, sacramental, character-level, naming) documented for completeness.  The workbook contains one sheet per novel and the following columns for each entry: \emph{\#}, \emph{McCarthy Passage}, \emph{KJV Verse(s)}, \emph{Connection Type}, \emph{Rating} (Strong/Moderate), \emph{Detection} (Pipeline / Gemini / Pipeline+Gemini / Missed), \emph{Attestation} (Attested / Unattested), \emph{Textual} (Yes / No), \emph{Allusion Type} (Allusion / Reference), \emph{Citation(s)}, \emph{Verified?}, \emph{Notes}, and source-text offsets (\emph{MC chunk}, \emph{MC passage start/end}, \emph{KJV chunk}, \emph{KJV passage start/end}).

Cell colours encode the Detection and Attestation columns consistently across all sheets:

\begin{itemize}
    \item \textbf{Detection}: Pipeline (light green), Pipeline+Gemini (medium green), Gemini (light blue), Missed (peach).
    \item \textbf{Attestation}: Attested (light lavender), Unattested (light yellow).
\end{itemize}

This colour scheme is intended to make the provenance and scholarly status of each finding visible at a glance when the workbook is read alongside the manuscript.

\end{document}